\documentclass[11pt,a4paper]{article}
\usepackage[hyperref]{naaclhlt2019}
\usepackage{times}
\usepackage{latexsym}
\usepackage{todonotes}
\usepackage{booktabs}
\usepackage{pgfplots}
\usepackage{siunitx}
\usepackage{amsbsy}
\usepackage{url}
\usepackage{multicol}
\usepackage{multirow}
\pgfplotsset{compat=1.14}

\aclfinalcopy 
\def\aclpaperid{28} 

\newcommand{\fr}{\boldsymbol}

\title{Publicly Available Clinical BERT Embeddings}

\author{
  Emily Alsentzer \\
  Harvard-MIT \\
  Cambridge, MA\\
  {\tt emilya@mit.edu} \\
  \And
  John R. Murphy \\
  MIT CSAIL \\
  Cambridge, MA \\
  {\tt jrmurphy@mit.edu} \\
  \And
  \hspace{-3mm}Willie Boag \\
  \hspace{-3mm}MIT CSAIL \\
  \hspace{-3mm}Cambridge, MA \\
  \hspace{-3mm}{\tt wboag@mit.edu} \\
  \And
  \hspace{-5mm}Wei-Hung Weng \\
  \hspace{-5mm}MIT CSAIL \\
  \hspace{-5mm}Cambridge, MA \\
  \hspace{-5mm}{\tt ckbjimmy@mit.edu} \\
  \AND
  Di Jin \\
  MIT CSAIL \\
  Cambridge, MA \\
  {\tt jindi15@mit.edu} \\
  \And
  Tristan Naumann \\
  Microsoft Research \\
  Redmond, WA\\
  {\tt tristan@microsoft.com} \\
  \And
  Matthew B. A. McDermott \\
  MIT CSAIL \\
  Cambridge, MA \\
  {\tt mmd@mit.edu} \\
}
  
\date{}

\hypersetup{draft}
\begin{document}
\maketitle

\begin{abstract}
    Contextual word embedding models such as ELMo~\cite{peters_deep_2018} and BERT~\cite{devlin_bert:_2018} have dramatically improved performance for many natural language processing (NLP) tasks in recent months. However, these models have been minimally explored on specialty corpora, such as clinical text; moreover, in the clinical domain, no publicly-available pre-trained BERT models yet exist. In this work, we address this need by exploring and releasing BERT models for clinical text: one for generic clinical text and another for discharge summaries specifically. We demonstrate that using a domain-specific model yields performance improvements on three common clinical NLP tasks as compared to nonspecific embeddings. These domain-specific models are not as performant on two clinical de-identification tasks, and argue that this is a natural consequence of the differences between de-identified source text and synthetically non de-identified task text.
\end{abstract}

\section{Introduction}
Natural language processing (NLP) has been shaken in recent months with the dramatic successes enabled by transfer learning and contextual word embedding models, such as ELMo \cite{peters_deep_2018}, ULMFiT \cite{howard_universal_2018}, and BERT \cite{devlin_bert:_2018}. 

These models have been primarily explored for general domain text, and, recently, biomedical text with BioBERT \cite{lee_biobert:_2019}. However, clinical narratives (e.g., physician notes) have known differences in linguistic characteristics from both general text and non-clinical biomedical text, motivating the need for specialized clinical BERT models.
  

In this work, we build and publicly release exactly such an embedding model.\footnote{\texttt{github.com/EmilyAlsentzer/clinicalBERT}} Furthermore, we demonstrate on several clinical NLP tasks the improvements this system offers over traditional BERT and BioBERT alike.
  
In particular, we make the following contributions:
\begin{enumerate}
  \item We train and publicly release BERT-Base and BioBERT-finetuned models trained on both all clinical notes and only discharge summaries.\footnote{Discharge summaries are commonly used in downstream tasks.}
  \item We demonstrate that using clinical specific contextual embeddings improves both upon general domain results and BioBERT results across 2 well established clinical NER tasks and one medical natural language inference task (i2b2 2010 \cite{uzuner_2010_2011}, i2b2 2012 \cite{sun_annotating_2013,sun_evaluating_2013}, and MedNLI \cite{romanov_lessons_2018}). On 2 de-identification (de-ID) tasks, i2b2 2006 \cite{uzuner_evaluating_2007} and i2b2 2014 \cite{stubbs_automated_2015,stubbs_annotating_2015}, general BERT and BioBERT outperform clinical BERT and we argue that fundamental facets of the de-ID context motivate this lack of performance.
\end{enumerate}

\section{Related Work}








\paragraph{Contextual Embeddings in General}
Traditional word-level vector representations, such as word2vec~\cite{mikolov2013distributed},  GloVe~\cite{pennington2014glove}, and fastText~\cite{bojanowski2017enriching}, express all possible meanings of a word as a single vector representation and cannot disambiguate the word senses based on the surrounding context. Over the last two years, ELMo~\cite{peters_deep_2018} and BERT~\cite{devlin_bert:_2018} present strong solutions that can provide contextualized word representations. By pre-training on a large text corpus as a language model, ELMo can create a context-sensitive embedding for each word in a given sentence, which will be fed into downstream tasks. Compared to ELMo, BERT is deeper and contains much more parameters, thus possessing greater representation power. More importantly, rather than simply providing word embeddings as features, BERT can be incorporated into a downstream task and gets fine-tuned as an integrated task-specific architecture.

BERT has, in general, been found to be superior to ELMo and far superior to non-contextual embeddings on a variety of tasks, including those in the clinical domain \cite{si_enhancing_2019}. For this reason, we only examine BERT here, rather than including ELMo or non-contextual embedding methods.

\paragraph{Contextual Clinical \& Biomedical Embeddings}
Several works have explored the utility of contextual models in the clinical and biomedical domains. BioBERT \cite{lee_biobert:_2019} trains a BERT model over a corpus of biomedical research articles sourced from PubMed\footnote{\url{https://www.ncbi.nlm.nih.gov/pubmed/}} article abstracts and PubMed Central\footnote{\url{https://www.ncbi.nlm.nih.gov/pmc/}} article full texts. They find the specificity offered by biomedical texts translated to improved performance on several biomedical NLP tasks, and fully release their pre-trained BERT model. 

On clinical text, \cite{khin_deep_2018} uses a general-domain pretrained ELMo model towards the task of clinical text de-identification, reporting near state-of-the-art performance on the i2b2 2014 task \cite{stubbs_annotating_2015,stubbs_automated_2015} and state of the art performance on several axes of the HIPAA PHI dataset.

Two works that we know of train contextual embedding models on clinical corpora.

\cite{zhu_clinical_2018} trains an ELMo model over a corpus of mixed clinical discharge summaries, clinical radiology notes and medically oriented wikipedia articles, then demonstrates improved performance on the i2b2 2010 task \cite{uzuner_2010_2011}. They release a pre-trained ELMo model along with their work, enabling further clinical NLP research to work with these powerful contextual embeddings.

\cite{si_enhancing_2019}, released in late February 2019, train a clinical note corpus BERT language model and uses complex task-specific models to yield improvements over both traditional embeddings and ELMo embeddings on the i2b2 2010 and 2012 tasks \cite{sun_evaluating_2013,sun_annotating_2013} and the SemEval 2014 task 7 \cite{pradhan_semeval-2014_2014} and 2015 task 14 \cite{elhadad_semeval-2015_nodate} tasks, establishing new state-of-the-art results on all four corpora. However, this work neither releases their embeddings for the larger community nor examines the performance opportunities offered by fine-tuning BioBERT with clinical text or by training note-type specific embedding models, as we do.

\section{Methods}
In this section, we first describe our clinical text dataset, the details of the BERT training procedure, and finally the specific tasks we examine.

\subsection{Data}
We use clinical text from the approximately 2 million notes in the MIMIC-III v1.4 database~\cite{johnson_mimic-iii_2016}. Details of our text pre-processing procedure can be found in Appendix~\ref{sec:MIMIC_notes}. Note that while some of our tasks use a small subset of MIMIC notes in their corpora, we do not try to filter these notes out of our BERT pre-training procedure. We expect the bias this induces is negligible given the relative sizes of the two corpora.

We train two varieties of BERT on MIMIC notes: Clinical BERT, which uses text from all note types, and Discharge Summary BERT, which uses only discharge summaries in an effort to tailor the corpus to downstream tasks (which often largely use discharge summaries).

Note that we train our clinical BERT instantiations on all notes of the appropriate type(s), without regard for whether or not any individual note appeared in any of the train/test sets for the various tasks we use (two of which use a small subset of MIMIC notes either partially or completely as their backing corpora). We feel this has a negligible impact given the dramatically larger size of the entire MIMIC corpus relative to the various task corpora.

\subsection{BERT Training}
In this work, we aim to provide the pre-trained embeddings as a community resource, rather than demonstrate technical novelty in the training procedure, and accordingly our BERT training procedure is completely standard. As such, we have relegated specifics of the training procedure to Appendix~\ref{sec:bert_training}.

We trained two BERT models on clinical text: 1) Clinical BERT, initialized from BERT-Base, and 2) Clinical BioBERT, initialized from BioBERT. For all downstream tasks, BERT models were allowed to be fine-tuned, then the output BERT embedding was passed through a single linear layer for classification, either at a per-token level for NER or de-ID tasks or applied to the sentinel ``begin sentence'' token for MedNLI.
Note that this is a substantially lower capacity model than, for example, the Bi-LSTM layer used in \cite{si_enhancing_2019}. This reduced capacity potentially limits performance on downstream tasks, but is in line with our goal of demonstrating the efficacy of clinical-specific embeddings and releasing a pre-trained BERT model for these embeddings. We did not experiment with more complex representations as our goal is not to necessarily surpass state-of-the-art performances on these tasks.

\paragraph{Computational Cost}
Pre-processing and training BERT on MIMIC notes took significant computational resources. We estimate that our entire embedding model procedure took roughly 17 - 18 days of computational runtime using a single GeForce GTX TITAN X 12 GB GPU (and significant CPU power and memory for pre-processing tasks). This is not including any time required to download or setup MIMIC or to train any final downstream tasks. 18 days of continuous runtime is a significant investment and may be beyond the reach of some labs or institutions. This is precisely why we believe that releasing our pre-trained model will be useful to the community.

\subsection{Tasks}

The Clinical BERT and Clinical BioBERT models were applied to the MedNLI natural language inference task \cite{romanov_lessons_2018} and four i2b2 named entity recognition (NER) tasks, all in IOB format~\cite{ramshaw_text_1995}: i2b2 2006 1B de-identification \cite{uzuner_evaluating_2007}, i2b2 2010 concept extraction \cite{uzuner_2010_2011}, i2b2 2012 entity extraction challenge \cite{sun_annotating_2013,sun_evaluating_2013}, i2b2 2014 7A de-identification challenge \cite{stubbs_annotating_2015, stubbs_automated_2015}. Details of the IOB format can be seen in the appendix, section~\ref{sec:iob_format}.
%
All task dataset sizes, evaluation metrics, and number of classes are shown in Table~\ref{tab:task_stats}.

Note that our two de-identification (de-ID) datasets present synthetically-masked PHI in their texts---e.g., they replace instances of real names, hospitals, etc., with synthetic, but consistent and realistic, names, hospitals, etc. As a result, they present significantly different text distributions than traditionally de-identified text (such as MIMIC notes) which will instead present sentinel ``PHI'' symbols at locations where PHI was removed.

\begin{table}[t]
    \centering
    \addtolength{\tabcolsep}{-3pt} 
    \begin{tabular}{llrrrr} \toprule
        \multirow{2}{*}{Dataset} &
        \multirow{2}{*}{Metric}  & 
        \multirow{2}{*}{Dim}     &
        \multicolumn{3}{c}{\# Sentences} \\
                  &          &    & Train & Dev  & Test \\ \midrule
        MedNLI    & Accuracy & 3  & 11232 & 1395 & 1422 \\ 
        i2b2 2006 & Exact F1 & 17 & 44392 & 5547 & 18095\\
        i2b2 2010 & Exact F1 & 7  & 14504 & 1809 & 27624\\
        i2b2 2012 & Exact F1 & 13 & 6624  & 820  & 5664 \\
        i2b2 2014 & Exact F1 & 43 & 45232 & 5648 & 32586\\
    \bottomrule \end{tabular}
    \caption{
        Task dataset evaluation metrics, output dimensionality, and train/dev/test dataset sizes (in number of sentences). Exact F1 requires that the text span and label be an exact match to be considered correct.
    }
    \label{tab:task_stats}
\end{table}


\begin{table*}[th!]
    \centering

    \begin{tabular}{lrrrrr} \toprule
        Model                      &   MedNLI    & i2b2 2006 & i2b2 2010 & i2b2 2012 & i2b2 2014 \\ \midrule
        BERT                       &  $77.6\%$   &$93.9$     & $83.5$    & $75.9$    & $92.8$          \\
        BioBERT                    &  $80.8\%$   &$\fr{94.8}$& $86.5$    & $78.9$    &   $\fr{93.0}$        \\
        Clinical BERT              &  $80.8\%$   &$91.5$     & $86.4$    & $78.5$    &  $92.6$        \\
        Discharge Summary BERT     &  $80.6\%$   &$91.9$     & $86.4$    & $78.4$    &   $92.8$         \\
        Bio+Clinical BERT          &$\fr{82.7\%}$&$94.7$     & $87.2$    &$\fr{78.9}$&    $92.5$      \\
        Bio+Discharge Summary BERT &  $\fr{82.7\%}$   &$94.8$     &$\fr{87.8}$& $78.9$    &   $92.7$        \\
    \bottomrule \end{tabular}


    \caption{
        Accuracy (MedNLI) and Exact F1 score (i2b2)  across various clinical NLP tasks.
    }
    \label{tab:embedding_performance}
\end{table*}
\begin{table*}[th!]
    \centering
    \scriptsize
    
    \addtolength{\tabcolsep}{-4pt} 
    \begin{tabular}{l|lll|lll|lll} \toprule
        \multirow{2}{*}{Model}   & \multicolumn{3}{|c|}{Disease}    & \multicolumn{3}{|c|}{Operations}  & \multicolumn{3}{|c}{Generic}  \\
                                 & Glucose  &Seizure  &Pneumonia    & Transfer  & Admitted  & Discharge & Beach& Newspaper  & Table     \\ \midrule
        \multirow{3}{*}{BioBERT} &insulin   &episode  &vaccine      &drainage   &admission  & admission & coast&news        &tables     \\
                                 &exhaustion&appetite &infection    &division   &sinking    & wave      & rock &official    &row        \\
                                 &dioxide   & attack  &plague       &transplant &hospital   & sight     & reef &industry    &dinner     \\ \midrule
        \multirow{3}{*}{Clinical}&potassium &headache &consolidation&transferred&admission  &disposition& shore&publication &scenario   \\
                                 &sodium    & stroke  &tuberculosis &admitted   &transferred& transfer  & ocean&organization&compilation\\
                                 &sugar     &agitation&infection    &arrival    &admit      &transferred& land &publicity   &technology \\
    \bottomrule \end{tabular}
    \caption{\normalsize Nearest neighbors for 3 sentinel words for each of 3 categories. In the Disease and operations categories, clinical BERT appears to show greater cohesion within the clinical domain than BioBERT, whereas for generic words, the methods do not differ much, as expected.}
    \label{tab:nearest_neighbors}
\end{table*}

\section{Results \& Discussions}
In this section, we will first describe quantitative comparisons of the various BERT models on the clinical NLP tasks we considered, and second describe qualitative evaluations of the differences between Clinical- and Bio- BERT.

\paragraph{Clinical NLP Tasks}
Full results are shown in Table~\ref{tab:embedding_performance}. On three of the five tasks (MedNLI, i2b2 2010, and i2b2 2012), clinically fine-tuned BioBERT shows improvements over BioBERT or general BERT. Notably, on MedNLI, clinical BERT actually yields a new state of the art, yielding a performance of 82.7\% accuracy as compared to the prior state of the art of 73.5\% \cite{romanov_lessons_2018} obtained via the InferSent model \cite{conneau_supervised_2017}. However, on our two de-ID tasks, i2b2 2006 and i2b2 2014, clinical BERT offers no improvements over Bio- or general BERT. This is actually not surprising, and is instead, we argue, a direct consequence of the nature of de-ID challenges.

De-ID challenge data presents a different data distribution than MIMIC text. In MIMIC, PHI is identified and replaced with sentinel PHI markers, whereas in the de-ID task, PHI is masked with synthetic, but realistic PHI. This data drift would be problematic for any embedding model, but will be especially damaging to contextual embedding models like BERT because the underlying sentence structure will have changed: in raw MIMIC, sentences with PHI will \emph{universally} have a sentinel PHI token. In contrast, in the de-ID corpus, all such sentences will have different synthetic masks, meaning that a canonical, nearly constant sentence structure present during BERT's training will be non-existent at task-time. For these reasons, we think it is sensible that clinical BERT is not successful on the de-ID corpora. Furthermore, this is a good example for the community given how prevalent the assumption is that contextual embedding models trained on task-like corpora will offer dramatic improvements.

Overall, we feel our results demonstrates the utility of using domain-specific contextual embeddings for non de-ID clinical NLP tasks. Additionally, on one task Discharge Summary BERT offers performance improvements over Clinical BERT, so it may be that adding greater specificity to the underlying corpus is helpful in some cases. We release both models with this work for public use.

\paragraph{Qualitative Embedding Comparisons}
Table~\ref{tab:nearest_neighbors} shows the nearest neighbors for 3 words each from 3 categories under BioBERT and Clinical BERT. These lists suggest that Clinical BERT retains greater cohesion around medical or clinic-operations relevant terms than does BioBERT. For example, the word ``Discharge'' is most closely associated with ``admission,'' ``wave,'' and ``sight'' under BioBERT, yet only the former seems relevant to clinical operations. In contrast, under Clinical BERT, the associated words all are meaningful in a clinical operations context.
\paragraph{Limitations \& Future Work}
This work has several notable limitations.
First, we do not experiment with any more advanced model architectures atop our embeddings. This likely hurts our performance.
Second, MIMIC only contains notes from the intensive care unit of a single healthcare institution (BIDMC). Differences in care practices across institutions are significant, and using notes from multiple institutions could offer significant gains. Lastly, our model shows no improvements for either de-ID task we explored. If our hypothesis is correct as to its cause, a possible solution could entail introducing synthetic de-ID into the source clinical text and using that as the source for de-ID tasks going forward.

\section{Conclusion}
In this work, we pretrain and release clinically oriented BERT models, some trained solely on clinical text, and others fine-tuned atop BioBERT. We find robust evidence that our clinical embeddings are superior to general domain or BioBERT specific embeddings for non de-ID tasks, and that using note-type specific corpora can induce further selective performance benefits. To the best of our knowledge, our work is the first to release clinically trained BERT models.
Our hope is that all clinical NLP researchers will be able to benefit from these embeddings without the necessity of the significant computational resources required to train these models over the MIMIC corpus.

\section{Acknowledgements}
This research was funded in part by grants from the
National Institutes of Health (NIH):
Harvard Medical School Biomedical Informatics and Data Science Research Training Grant T15LM007092 (Co-PIs: Alexa T. McCray, PhD and Nils Gehlenborg, PhD),
National Institute of Mental Health (NIMH) grant P50-MH106933,
National Human Genome Research Institute (NHGRI) grant U54-HG007963, and National Science Foundation Graduate Research Fellowship Program (NSF GRFP) under Grant No. 1122374. Additional funding was received from the MIT UROP program.




\clearpage

\bibliography{bibliography_di,bibliography}
\bibliographystyle{acl_natbib}

\clearpage

\appendix

\section{MIMIC Notes}
\label{sec:MIMIC_notes}
\begin{figure}[h]
    \centering
    \label{landuse}
    \begin{tikzpicture}
    \begin{axis}[
    ybar=0cm,
    axis x line*=bottom,
    axis y line*=left,
    height=5cm, width=\linewidth,
    ylabel={Note Count},
    symbolic x coords={Nursing/other,Radiology,Nursing,ECG,Physician,Discharge Summary,Other},
    xtick={Nursing/other,Radiology,Nursing,ECG,Physician,Discharge Summary,Other},
    x tick label style={rotate=60, anchor=east, align=left}
    ]
    \addplot[blue,fill] coordinates {(Nursing/other,822497) (Radiology,522279)(Nursing,223556)(ECG,209051)(Physician, 141624)(Discharge Summary,59652)(Other,104521) };
    \end{axis}
    \end{tikzpicture}
    \caption{Relative prevalence of MIMIC notes types.}
    \label{fig:notes_stats}
\end{figure}
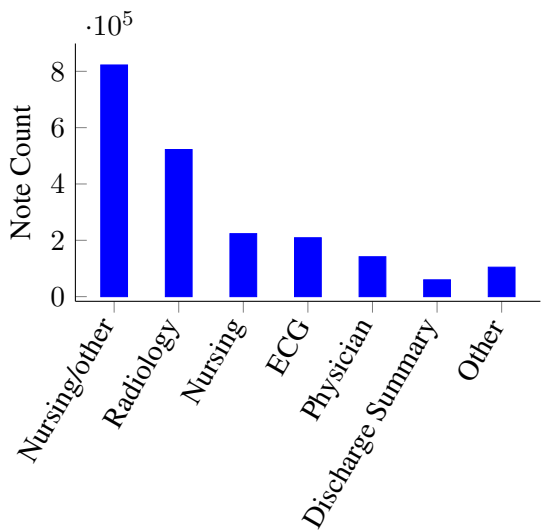

MIMIC notes are distributed among 15 note types (Figure~\ref{fig:notes_stats}). Many note types are semi-structured, with section headers separating free text paragraphs. To process these notes, we split all notes into sections, then used Scispacy \cite{neumann_scispacy:_2019} (specifically, the \texttt{en\_core\_sci\_md} tokenizer) to perform sentence extraction. The sentences are input into the BERT-Base and BioBERT models for additional pre-training on clinical text.
 
\section{BERT Training Details}
\label{sec:bert_training}

For all pre-training experiments, we leverage the tensorflow implementation of BERT~\cite{devlin_bert:_2018}.\footnote{https://github.com/google-research/bert}

\subsection{Pre-training}
We used a batch size of 32, a maximum sequence length of 128, and a learning rate of $5 \cdot 10^{-5}$ for pre-training our models. Models were trained for 150,000 steps. We experimented with models pre-trained for 300,000 steps, but we found no significant differences in downstream task performance with these models. The dup factor for duplicating input data with different masks was set to 5. All other default parameters were used (specifically, masked language model probability = 0.15 and max predictions per sequence = 20). 

\subsection{Fine-tuning}
For all downstream tasks, we explored the following hyperparameters: learning rate  $\in \{2 \cdot 10^{-5}, 3 \cdot 10^{-5}, 5 \cdot 10^{-5} \}$, 
batch size $\in \{16, 32\}$, and epochs $\in \{3, 4\}$. For the NER tasks, we also tried epoch $\in \{2\}$. The maximum sequence length was 150 across all tasks. Due to time constraints, only 2 epochs were run for the i2b2 2014 task.

\section{IOB Format}
\label{sec:iob_format}
The IOB (Inside-Outside-Beginning) format~\cite{ramshaw_text_1995} is a method of encoding span-based NER tasks to add more granularity to the label space over span positions, specifically re-classifying each class as having three subclasses:
\begin{description}
    \item[Inside (I-)] This label is used to specify words \emph{within a span} for this class.
    \item[Outside (O)] This label is used to specify words \emph{outside any span} for this class. This label will be shared across all classes and will replace the ``no class'' label applied to extraneous words.
    \item[Beginning (B-)] This label is used to specify words at the \emph{beginning of a span} for this class.
\end{description}

For example, if the input text, with span labels is given as
\\\textit{``The patient is very sick.''}\\
with NER labels
\\\textit{``Null Null Null Problem Problem''}\\
we could convert this into IOB format via
\\\textit{``O O O B-Problem I-Problem''}

\end{document}